# Dealing with the Fuzziness of Human Reasoning*

**Michael Gr. Voskoglou** [1], **Igor Ya. Subbotin** [2]

[1] Professor of Mathematical Sciences
School of Technological Applications
Graduate Technological Educational Institute, Patras, Greece
e-mail: mvosk@hol.gr , voskoglou@teipat.gr

[2] Professor of Mathematics
College of Letters and Sciences,
National University, Los Angeles, California, USA
e-mail: isubboti@nu.edu

## Abstract

***Reasoning, the most important human brain operation, is characterized by a degree of fuzziness. In the present paper we construct a fuzzy model for the reasoning process giving through the calculation of probabilities and possibilities of all possible individuals' profiles a quantitative/qualitative view of their behaviour during the above process. In this model the main stages of human reasoning (imagination, visualisation and generation of ideas) are represented as fuzzy subsets of a set of linguistic labels characterizing a person's performance in each stage. Further, using the coordinates of the centre of gravity of the graph of the corresponding membership function we develop a method of measuring the reasoning skills of a group of individuals. We also present a number of classroom experiments with student groups' of T. E. I. of Patras, Greece, illustrating our results in practice.***

## 1. Introduction

Early humans were thinking for their daily activities in an entirely uncertain environment, where concepts derived from frequent observations and experience. However, throughout the centuries human thinking had support from scripts, drawings, logic and finally mathematical calculations. The steps necessary in a complete thinking process include *understanding*, *explanation* and *reasoning*.

---

* The main ideas of this paper were presented by the first of the authors in the 6th International Conference on Information Technology (IICIT 13), Amman, Jordan, May 8-10, 2013.

Having a problem at hand, once the collection of the related linguistic information from the environment is completed, the human inquiry expands the field of understanding along different directions. In the mind each item concerning the phenomenon under investigation is labeled by a word or a set of words (statements, propositions). This is equivalent with the categorization of the objects into different classes. In this way the natural, environmental or engineering reality is divided into fragments and categories, which are fundamental ingredients in classification, analysis and deduction of conclusions.

Conscious direction of attention towards an external object causes the object to be received by mind in sequence to perception, experience, feeling, understanding, explaining, knowing, and finally acting for meaningful description and analytical solution.

Reasoning is the most important human brain operation that leads to creative methodologies, algorithms and deductions giving way to sustainable research and development. The main stages of the human reasoning for reaching to a solution of any problem in general involve *imagination, visualization* and *idea generations* ([2; p. 340] or [3; section 4]).

For any external object, whether it exists materialistically or not, human beings try to imagine its properties in their minds. This gives them the power of initializing their individual thinking domain with whole freedom in any direction. Imagination includes the setting up of a suitable hypothesis or a set of logical rules for the problem at hand.

The visualization stage is to defend the representative hypothesis and logical propositions. Humans typically use a variety of representations to defend their hypotheses including algorithms, graphs, diagrams, charts, figures, etc. In particular, the *geometric configuration* of the objects appearing through imagination is the most common among these representations. In fact, after an object comes into existence vaguely in mind, it is necessary to know its shape, which is related to geometry. It is essential that the geometric configuration of the phenomenon must be visualized in mind in some way, even though it may be a simplification under a set of assumptions.

On the basis of their hypotheses the individuals generate relevant ideas. The ideas begin to crystallize and they are expressed verbally by a native language to other individuals to get their criticisms, comments, suggestions and support for the betterment of the mental thinking and scientific achievement. Finally, all the conclusions must be expressed in a language, which can then be converted into universally used symbolic logic based on the principles of mathematics. We emphasize that, whatever are the means of reasoning, the scientific arguments are expressed verbally prior to any symbolic and mathematical abstractions.

In concluding, three are the essential steps in the human reasoning for scientific and technological achievements: The *perceptions* (feelings through imagination), the *sketches* (geometry, design) and the *ideas.* The perception part is very significant, because it provides complete freedom of thinking without expressing it to others, who can restrict the activity. The subjectivity is the main characteristic of the above part, but as one enters the sketch domain the subjectivities decrease and at the final stage, since the ideas are exposed to other individuals, the objectivity overrules becoming at least logical.

It is possible to state that with Newtonian classical physics science entered almost entirely a deterministic world, where uncertainty was not even accounted among the scientific knowledge. However, nowadays uncertainty appears in almost all branches of science and many scientific deterministic foundations of the past became uncertain with fuzzy ingredients. Among such conceptions are quantum physics, fractal geometry, chaos theory and fuzzy inference systems.

With the advancement of numerical uncertainty techniques, such as probability, statistics and stochastic principles, scientific progress in quantitative aspects had a rapid development, but still leaving aside the qualitative sources of knowledge and information, which can be tackled by the fuzzy logic principles only. Zadeh, the instructor of fuzzy logic through the fuzzy sets theory [10], states: "As the complexity of a system increases, our ability to make precise and yet significant statements about its behavior diminishes, until a threshold is reached beyond which precision and significance (or relevance) become almost mutually exclusive characteristics" [11].

This paper proposes the use of fuzzy logic for a more realistic description of the process of human reasoning. The text is organized as follows: In section 2 we develop a fuzzy model representing the reasoning process, while in section 3 we obtain, in terms of the above model, a measure of the reasoning abilities of a group of individuals. The application of our results in practice is illustrated in section 4, were we present classroom experiments performed with student groups. Finally, in section 5 we state our final conclusions and we discuss the future perspectives of our research.

## 2. The fuzzy model

The stages of the reasoning process presented above are helpful in understanding the individuals' *'ideal behaviour'* during the process. However, things in real situations are usually not happening like that, since human cognition utilizes in general concepts that are inherently graded and therefore fuzzy. This fact gave us the impulsion to introduce principles of fuzzy sets theory in order to describe in a more effective way the process of scientific reasoning. For general facts on fuzzy sets we refer freely to the book [1].

For the development of our fuzzy model for the reasoning process we consider a group of *n* people, n≥2, working (each one individually) on the same problem. Denote by $S_1$, $S_2$ and $S_3$ respectively the stages of imagination, visualization and ideas generation of the reasoning process. Denote also by *a, b, c, d,* and *e* the linguistic labels of very low, low, intermediate, high and very high success respectively of a person in each of the $S_i$*'s*. Set *U = {a, b, c, d, e}*. We are going to attach to each stage $S_i$ of the reasoning process, i=1, 2, 3 , a fuzzy subset, $A_i$ of *U*. For this, if $n_{ia}$, $n_{ib}$, $n_{ic}$, $n_{id}$ and $n_{ie}$ denote the number of individuals that faced very low, low, intermediate, high and very high success at stage $S_i$ respectively, i=*1,2,3,* we define the *membership function* $m_{Ai}$ for each *x* in *U*, as follows:

$$m_{A_i}(x) = \begin{cases} 1, & if \ \frac{4n}{5} < n_{ix} \le n \\ 0{,}75, & if \ \frac{3n}{5} < n_{ix} \le \frac{4n}{5} \\ 0{,}5, & if \ \frac{2n}{5} < n_{ix} \le \frac{3n}{5} \\ 0{,}25, & if \ \frac{n}{5} < n_{ix} \le \frac{2n}{5} \\ 0, & if \ 0 \le n_{ix} \le \frac{n}{5} \end{cases}$$

In fact, if one wanted to apply probabilistic standards in measuring the degree of the individuals' success at each stage of the process, then he/she should use the relative frequencies $\frac{n_{ix}}{n}$. Nevertheless, such an action would be highly questionable, since the $n_{ix}$*'s* are obtained with respect to the linguist labels of U, which are fuzzy expressions by themselves. Therefore the application of a fuzzy approach by using membership degrees instead of probabilities seems to be more suitable for this case. But, as it is well known, the membership function needed for such purposes is usually defined empirically in terms of logical or/and statistical data. In our case the above definition of $m_{A_i}$ seems to be compatible with the common logic.

Then the fuzzy subset $A_i$ of *U* corresponding to $S_i$ has the form: $A_i = \{(x, m_{Ai}(x)): x \in U\}$, *i=1, 2, 3.*

In order to represent all possible individuals' *profiles (overall states)* during the reasoning process we consider a *fuzzy relation*, say *R*, in $U^3$ of the form: $R = \{(s, m_R(s)): s=(x, y, z) \in U^3\}$.

For determining properly the membership function $m_R$ we give the following definition:

*A profile s=(x, y, z), with x, y, z in U, is said to be well ordered if x corresponds to a degree of success equal or greater than y and y corresponds to a degree of success equal or greater than z.*

For example, *(c, c, a)* is a well ordered profile, while *(b, a, c)* is not.

We define now the *membership degree* of a profile *s* to be $m_R(s) = m_{A_1}(x)m_{A_2}(y)m_{A_3}(z)$ if *s* is well ordered, and *0* otherwise.

In fact, if for example the profile *(b, a, c)* possessed a nonzero membership degree, how it could be possible for a person, who has failed at the visualization stage, to perform satisfactorily at the stage of the ideas generation?

Next, for reasons of brevity, we shall write $m_s$ instead of $m_R(s)$. Then the *probability* $p_s$ of the profile s is defined in a way analogous to crisp data, i.e. by $P_s = \frac{m_s}{\sum_{s\in U^3} m_s}$ .

We define also the *possibility* $r_s$ of *s* to be $r_s = \frac{m_s}{\max\{m_s\}}$ , where $max\{m_s\}$ denotes the maximal value of $m_s$ , for all *s* in $U^3$. In other words the possibility of *s* expresses the "relative membership degree" of *s* with respect to $max\{m_s\}$.

Assume further that one wants to study the *combined results* of behaviour of *k* different groups of people, $k\geq 2$, during the reasoning process. For this, we introduce the *fuzzy variables* $A_1(t)$, $A_2(t)$ and $A_3(t)$ with $t=1, 2,..., k$. The values of these variables represent fuzzy subsets of *U* corresponding to the stages of the reasoning process for each of the *k* groups; e.g. $A_1(2)$ represents the fuzzy subset of *U* corresponding to the stage of imagination for the second group ($t=2$).

Obviously, in order to measure the degree of evidence of the combined results of the *k* groups, it is necessary to define the probability *p(s)* and the possibility *r(s)* of each profile *s* with respect to the membership degrees of *s* for all groups. For this reason we introduce the *pseudo-frequencies* $f(s) = \sum_{t=1}^{k} m_s(t)$ and we define the probability and possibility of a profile *s* by $p(s) = \frac{f(s)}{\sum_{s\in U^3} f(s)}$ and $r(s) = \frac{f(s)}{\max\{f(s)\}}$ respectively, where $max\{f(s)\}$ denotes the maximal pseudo-frequency.

Obviously the same method could be applied when one wants to study the combined results of behaviour of a group during *k* different reasoning situations.

The above model gives, through the calculation of probabilities and possibilities of all individuals' profiles, a quantitative/qualitative view of their realistic performance at all stages of the reasoning process.

## 3. Measuring scientific reasoning skills

There are *natural* and *human-designed* real systems. In contrast to the former, which may not have an apparent objective, the latter are made with purposes that are achieved by the delivery of outputs. Their parts must be related, i.e. they must be designed to work as a coherent entity. The most important part of a human-designed system's study is probably the assessment, through the model representing it, of its performance. In fact, this could help the system's designer to make all the necessary modifications/improvements to the system's structure in order to increase its effectiveness.

The amount of information obtained by an action can be measured by the reduction of uncertainty resulting from this action. Accordingly a system's uncertainty is connected to its capacity in obtaining relevant information. Therefore a measure of uncertainty could be adopted as a measure of a system's effectiveness in solving related problems. Based on this fact, we have used in earlier papers the total possibilistic uncertainty, as well as the Shannon's entropy (total probabilistic uncertainty) - properly adapted for use in a fuzzy environment - for measuring the effectiveness of several systems in the areas of Education, of Artificial Intelligence and of Management (e.g. Problem Solving, Learning, Case-Based Reasoning, evaluation of the fuzzy data of a market's research, etc); see the book [8] and its references and [9].

In this paper and in terms of the fuzzy model developed above we shall introduce another approach for measuring the groups' reasoning capacities, known as the *'centroid method'*. According to this method the centre of gravity of the graph of the membership function involved provides an alternative measure of the system's performance. The application of the 'centroid method' in practice is simple and evident and, in contrast to the measures of uncertainty, needs no complicated calculations in its final step. The techniques that we shall apply here have been also used earlier in [5], [6], [7], etc..

Given a fuzzy subset $A = \{(x, m(x)): x \in U\}$ of the universal set $U$ of the discourse with membership function $m: U \rightarrow [0, 1]$, we correspond to each $x \in U$ an interval of values from a prefixed numerical distribution, which actually means that we replace $U$ with a set of real intervals. Then, we construct the graph $F$ of the membership function $y=m(x)$.There is a commonly used in fuzzy logic approach to measure performance with the pair of numbers ($x_c$, $y_c$) as the coordinates of the *centre of gravity*, say $F_c$, of the graph *F*, which we can calculate using the following well-known formulas:

$$x_c = \frac{\iint_F x dx dy}{\iint_F dx dy}, \quad y_c = \frac{\iint_F y dx dy}{\iint_F dx dy} \qquad (1)$$

Concerning the reasoning process, we characterize an individual's performance as very low (a) if $x \in [0, 1)$, as low (b) if $x \in [1, 2)$, as intermediate (c) if $x \in [2, 3)$, as high (d) if $x \in [3, 4)$ and as very high (e) if $x \in [4, 5]$ respectively. Therefore in this case the graph F of the corresponding fuzzy subset of U is the bar graph of Figure 1 consisting of five rectangles, say $F_i$, i=1,2,3, 4, 5 , whose sides lying on the x axis have length 1.

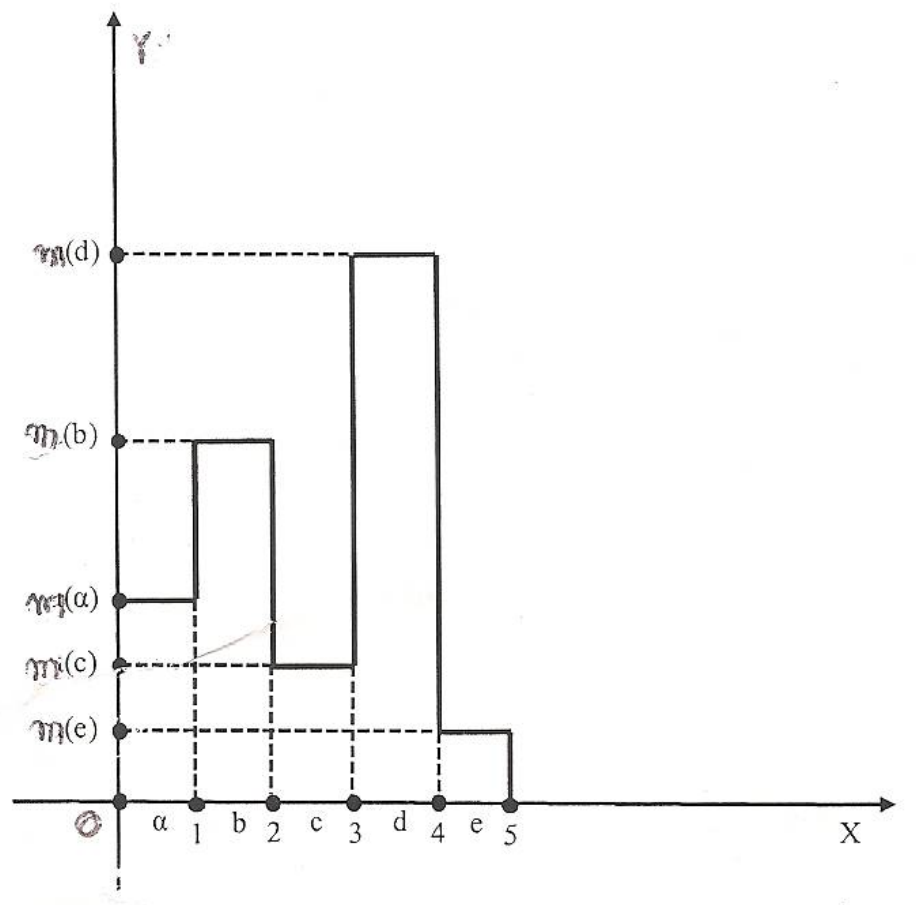

Figure 1: Bar graphical data representation

In this case $\iint_F dx dy$ is the area of F which is equal to $\sum_{i=1}^{5} y_i$. Also $\iint_F x dx dy$ $= \sum_{i=1}^{5} \iint_{F_i} x dx dy = \sum_{i=1}^{5} \int_0^{y_i} dy \int_{i-1}^{i} x dx = \sum_{i=1}^{5} y_i \int_{i-1}^{i} x dx = \frac{1}{2} \sum_{i=1}^{5} (2i-1) y_i$, and $\iint_F y dx dy = \sum_{i=1}^{5} \iint_{F_i} y dx dy = \sum_{i=1}^{5} \int_0^{y_i} y dy \int_{i-1}^{i} dx =$ $\sum_{i=1}^{n} \int_0^{y_i} y dy = \frac{1}{2} \sum_{i=1}^{n} y_i^2$. Therefore formulas (1) are transformed into the following form:

$$x_c = \frac{1}{2}\left( \frac{y_1 + 3y_2 + 5y_3 + 7y_4 + 9y_5}{y_1 + y_2 + y_3 + y_4 + y_5} \right),$$

$$y_c = \frac{1}{2}\left( \frac{y_1^2 + y_2^2 + y_3^2 + y_4^2 + y_5^2}{y_1 + y_2 + y_3 + y_4 + y_5} \right). \qquad (2)$$

Normalizing our fuzzy data by dividing each m(x), $x \in U$, with the sum of all membership degrees we can assume without loss of the generality that $y_1+y_2+y_3+y_4+y_5 = 1$. Therefore we can write:

$$x_c = \frac{1}{2}\left(y_1 + 3y_2 + 5y_3 + 7y_4 + 9y_5\right),$$
$$y_c = \frac{1}{2}\left(y_1^2 + y_2^2 + y_3^2 + y_4^2 + y_5^2\right) \quad (3)$$

with $y_i = \frac{m(x_i)}{\sum_{x \in U} m(x)}$..

But $0 \leq (y_1-y_2)^2 = y_1^2+y_2^2-2y_1y_2$ , therefore $y_1^2+y_2^2 \geq 2y_1y_2$ ,with the equality holding if, and only if, $y_1=y_2$.

In the same way one finds that $y_1^2+y_3^2 \geq 2y_1y_3$, and so on. Hence it is easy to check that

$(y_1+y_2+y_3+y_4+y_5)^2 \leq 5(y_1^2+y_2^2+y_3^2+y_4^2+y_5^2)$, with the equality holding if, and only if $y_1=y_2=y_3=y_4=y_5$.

But $y_1+y_2+y_3+y_4+y_5 =1$, therefore $1 \leq 5(y_1^2+y_2^2+y_3^2+y_4^2+y_5^2)$ (4), with the equality holding if, and only if $y_1=y_2=y_3=y_4=y_5=\frac{1}{5}$ .

Then the first of formulas (3) gives that $x_c = \frac{5}{2}$. Further, combining the inequality (4) with the second of formulas (3) one finds that $1 \leq 10y_c$, or $y_c \geq \frac{1}{10}$ Therefore the unique minimum for $y_c$ corresponds to the centre of mass $F_m(\frac{5}{2}, \frac{1}{10})$.

The ideal case is when $y_1=y_2=y_3=y_4=0$ and $y_5=1$. Then from formulas (3) we get that $x_c = \frac{9}{2}$ and $y_c = \frac{1}{2}$.Therefore the centre of mass in this case is the point $F_i(\frac{9}{2}, \frac{1}{2})$.

On the other hand the worst case is when $y_1=1$ and $y_2=y_3=y_4= y_5=0$. Then for formulas (3) we find that the centre of mass is the point $F_w(\frac{1}{2}, \frac{1}{2})$.

Therefore the "area" where the centre of mass $F_c$ lies is represented by the triangle $F_w F_m F_i$ of Figure 2.

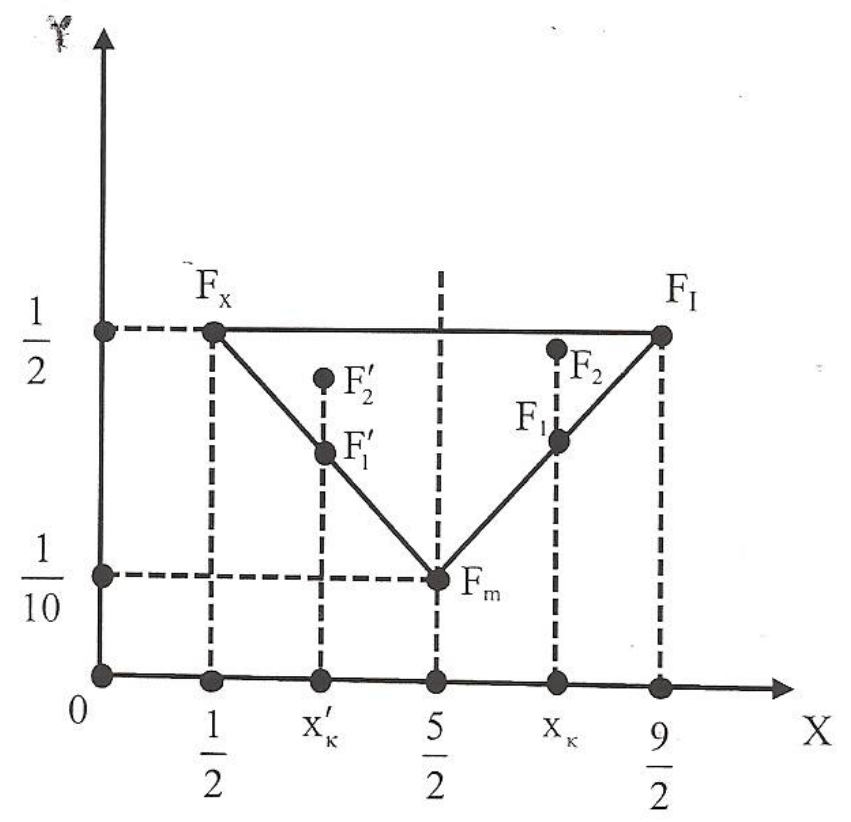

Figure 2: Graphical representation of the "area" of the centre of mass

Then from elementary geometric considerations it follows that the greater is the value of $x_c$ the better is the group's performance. Also, for two groups with the same $x_c \geq 2,5$, the group having the centre of mass which is situated closer to $F_i$ is the group with the higher $y_c$; and for two groups with the same $x_c < 2.5$ the group having the centre of mass which is situated farther to $F_w$ is the group with the lower $y_c$. Based on the above considerations it is logical to formulate our criterion for comparing the groups' performances in the following form:

- *Among two or more groups the group with the biggest $x_c$ performs better.*
- *If two or more groups have the same $x_c \geq 2.5$, then the group with the higher $y_c$ performs better.*
- *If two or more groups have the same $x_c < 2.5$, then the group with the lower $y_c$ performs better.*

# 4. Classroom experiments

In order to illustrate the use of our fuzzy model developed above in practice, we performed recently the following two experiments at the Graduate Technological Educational Institute (T. E. I.) of Patras, Greece.

In the first experiment our subjects were 35 students of the School of Technological Applications, i.e. future engineers. A few days before the experiment an analysis of the scientific reasoning process (see introduction) was presented to the students in a two hours lecture, followed by a number of suitable examples. The following two problems with their relevant analyses were included among these examples:

*Problem 1:* We want to construct a channel to run water by folding across its longer side the two edges of an orthogonal metallic leaf having sides of length *20 cm* and *32 cm,* in such a way that they will be perpendicular to the other parts of the leaf. Assuming that the flow of the water is constant, how we can run the maximum possible quantity of the water?

*Analysis of the problem:*

IMAGINATION: The basic thing to realize is that the quantity of water to run through the channel depends on the area of the vertical cut of the channel.

VISUALIZATION (geometric configuration): Folding the two edges of the metallic leaf by length x across its longer side the vertical cut of the constructed channel is an orthogonal with sides x and 32-2x (Figure 3).

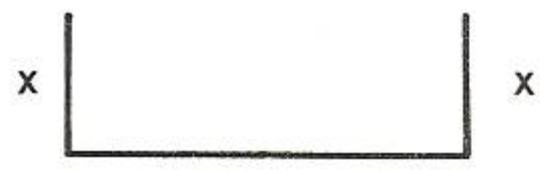

Figure 3: The vertical cut of the channel

IDEAS GENERATION: The area of the orthogonal has to be maximized.

The above idea leads to the following mathematical manipulation:

The area is equal to $E(x) = x(32\text{-}2x) = 32x\text{-}2x^2$. Taking the derivative $E'(x)$ the equation $E'(x) = 32\text{-}4x = 0$ gives that $x = 8\ cm$. But $E''(x) = -4 < 0$, therefore $E(8) = 128\ cm^2$ is the maximum possible quantity of water to run through the channel.

*Problem 2:* The rate of increase of the population of a country is analogous to the number of its inhabitants. If the population is doubled in 50 years, in how many years it will be tripled?

*Analysis of the problem:*

IMAGINATION: The key concepts involved in the statement of this problem are the ‘analogy’ and the ‘rate of increase’ of the population. Therefore the crucial action for the solution of the problem is to establish the relation connecting these two concepts.

VISUALIZATION: The population $P$ of the country is obviously a function of the time $t$, say $P = P(t)$. In observing the increase of the population we must consider a starting point, where $t = 0$.

IDEAS GENERATION: The rate of increase of the population is expressed by the derivative $P'(t)$ and the existing analogy is expressed by $P'(t) = k\,P(t)$, with $k$ a non negative integer. Therefore the solution of the problem is based on the solution of the above differential equation.

This suggests the following mathematical manipulation:

Separating the variables we can write $\frac{dP(t)}{P(t)} = kdt$, or $\int \frac{dP(t)}{P(t)} = k\int dt$. Thus $\ln P(t) = kt + \ln c = \ln e^{kt} + \ln c = \ln (c\, e^{kt})$, or $P(t) = c\, e^{kt}$. For $t = 0$ we find that $P(0) = P_0 = c$ and therefore we get that $P(t) = P_0 e^{kt}$ (1) .

Further, according to the problem's statement, we have that $P(50) = 2P_0$, or $P_0 e^{50k} = 2P_0 \Rightarrow 50k = \ln 2$, or $k = \frac{\ln 2}{50}$. Therefore (1) finally gives that $P(t) = P_0 e^{\frac{t \ln 2}{50}}$.

If the population will be tripled after $x$ years, then we'll have $3P_0 = P(x) = P_0 e^{\frac{x \ln 2}{50}}$, or $3 = e^{\frac{x \ln 2}{50}}$, which gives that $x = 50\frac{\ln 3}{\ln 2} \approx 79$ years. –

In performing the experiment the following problem was given for solution to the students (time allowed 20 minutes):

*Problem 3:* Among all cylinders having a total surface of $180\pi$ $m^2$, which one has the maximal volume?

Before starting the experiment we gave the proper instructions to students emphasizing, among the others, that we were interested for all their efforts (successful or not) during the reasoning process, and therefore they must keep records on their papers for all of them, at all stages of the process. In particular, we asked them to provide an analysis of the solution of the given problem analogous to the analyses presented for the above two examples.

Ranking the students' papers by using the scale applied in section 3 (before the construction of Figure 1) we found that 15, 12 and 8 students had intermediate, high and very high success respectively at stage $S_1$ of imagination. Therefore we obtained that $n_{1a}=n_{1b}=0$, $n_{1c}=15$, $n_{1d}=12$ and $n_{1e}=8$. Thus, by the definition of the corresponding membership function given in the second section, $S_1$ is represented by a fuzzy subset of $U$ of the form: $A_1 = \{(a,0),(b,0),(c, 0.5),(d, 0.25),(e,0..25)$.

In the same way we represented the stages $S_2$ and $S_3$ as fuzzy sets in $U$ by
$A_2 = \{(a,0),(b,0),(c, 0.5),(d, 0.25),(e,0)\}$ and $A_3 = \{(a, 0.25),(b, 0.25),(c, 0.25),(d,0),(e,0)\}$ respectively.

Next we calculated the membership degrees of the $5^3$ (ordered samples with replacement of 3 objects taken from 5) in total possible students' profiles as it is described in the second section (see column of $m_s(1)$ in Table 1). For example, for the profile $s=(c, c, a)$ one finds that $m_s = m_{A_1}(c).\ m_{A_2}(c).\ m_{A_3}(a) = 0.5 \times 0 .5 \times 0.25 = 0.06225$.

Further, from the values of the column of $m_s(1)$ it turns out that the maximal membership degree of students' profiles is *0.06225*. Therefore the possibility of each *s* in $U^3$ is given by $r_s = \frac{m_s}{0.06225}$.

One, in order to be able to make the corresponding comparisons, could also calculate the probabilities of the students' profiles using the formula for $p_s$ given in section 2. However notice that, according to Shackle [4] and many others after him, human reasoning is better presented by possibility rather than by probability theory. Therefore, adopting the above view, we considered that the calculation of the probabilities is not necessary.

Table 1: Profiles with non zero membership degrees

| $A_1$ | $A_2$ | $A_3$ | $m_s(1)$ | $r_s(1)$ | $m_s(2)$ | $r_s(2)$ | $f(s)$ | $r(s)$ |
|---|---|---|---|---|---|---|---|---|
| b | b | b | 0 | 0 | 0.016 | 0.258 | 0.016 | 0.129 |
| b | b | a | 0 | 0 | 0.016 | 0.258 | 0.016 | 0.129 |
| b | a | a | 0 | 0 | 0.016 | 0.258 | 0.016 | 0.129 |
| c | c | c | 0.062 | 1 | 0.062 | 1 | 0.124 | 1 |
| c | c | a | 0.062 | 1 | 0.062 | 1 | 0.124 | 1 |
| c | c | b | 0 | 0 | 0.031 | 0.5 | 0.031 | 0.25 |
| c | a | a | 0 | 0 | 0.031 | 0.5 | 0.031 | 0.25 |
| c | b | a | 0 | 0 | 0.031 | 0.5 | 0.031 | 0.25 |
| c | b | b | 0 | 0 | 0.031 | 0.5 | 0.031 | 0.25 |
| d | d | a | 0.016 | 0.258 | 0 | 0 | 0.016 | 0.129 |
| d | d | b | 0.016 | 0.258 | 0 | 0 | 0.016 | 0.129 |
| d | d | c | 0.016 | 0.258 | 0 | 0 | 0.016 | 0.129 |
| d | a | a | 0 | 0 | 0.016 | 0.258 | 0.016 | 0.129 |
| d | b | a | 0 | 0 | 0.016 | 0.258 | 0.016 | 0.129 |
| d | b | b | 0 | 0 | 0.016 | 0.258 | 0.016 | 0.129 |
| d | c | a | 0.031 | 0.5 | 0.031 | 0.5 | 0.062 | 0.5 |
| d | c | b | 0.031 | 0.5 | 0.031 | 0.5 | 0.062 | 0.5 |
| d | c | c | 0.031 | 0.5 | 0.031 | 0.5 | 0.062 | 0.5 |
| e | c | a | 0.031 | 0.5 | 0 | 0 | 0.031 | 0.25 |
| e | c | b | 0.031 | 0.5 | 0 | 0 | 0.031 | 0.25 |
| e | c | c | 0.031 | 0.5 | 0 | 0 | 0.031 | 0.25 |
| e | d | a | 0.016 | 0.258 | 0 | 0 | 0.016 | 0.129 |
| e | d | b | 0.016 | 0.258 | 0 | 0 | 0.016 | 0.129 |
| e | d | c | 0.016 | 0.258 | 0 | 0 | 0.016 | 0.129 |

The outcomes of Table 1 were obtained with accuracy up to the third decimal point.

A few days later we performed the same experiment with a group of 50 students from the School of Management and Economics. Working as in the first experiment we found that:
$A_1$=*{(a, 0),(b, 0.25),(c, 0.5),(d, 0.25),(e, 0)}*, $A_2$=*{(a, 0.25),(b, 0.25),(c, 0.5),(d, 0),(e, 0)}* and $A_3$=*{(a, 0.25),(b, 0.25),(c, 0.25),(d, 0),(e, 0)}*.

Then we calculated the membership degrees of all possible profiles of the student group (column of $m_s$ (2) in Table 1). Further, since the maximal membership degree is again 0.06225, the possibility of each *s* is given by the same formula as for the first group. The values of the possibilities of all profiles are given in column of $r_s(2)$ of Table 1.

Finally, in order to study the combined results of the two groups' performance we calculated the pseudo-frequencies $f(s) = m_s(1)+m_s(2)$ and the combined possibilities of all profiles (see the last two columns of Table 1) as it has been described in section 2 of the present paper.

Next, in order to compare the two groups' performance by the 'centroid method', let us denote by $A_{ij}$ the fuzzy subset of U attached to the stage $S_j$ , j=1,2,3 , of the reasoning process with respect to the student group i, i=1,2.

At the first stage of imagination we have $A_{11}$ = *{(a, 0),(b, 0),(c, 0.5),(d, 0.25),(e, 0.25),*

$A_{21}$= *{(a, 0),(b, 0.25),(c, 0.5),(d , 0.25),(e, 0)}* and respectively

$x_{c11} = \frac{1}{2}(5 \text{ x } 0.5 + 7 \text{ x } 0.25 + 9 \text{ x } 0.25) = 3.25$, $x_{c21} = \frac{1}{2}(3 \text{ x } 0.25 + 5 \text{ x } 0.5 + 7 \text{ x } 0.25) = 2.25$

Thus, by our criterion the first group demonstrates better performance.

At the second stage of visualization we have: $A_{12}$ = {(a, 0),(b, 0),(c, 0.5),(d, 0.25),(e, 0)} and $A_{22}$={(a, 0.25),(b, 0.25),(c, 0.5),(d, 0),(e, 0)}.

Normalizing the membership degrees in the first of the above fuzzy subsets of U (0.5 : 0,.75 ≈ 0.67 and 0.25 : 0.75 ≈ 0.33) we get $A_{12}$ = {(a, 0),(b, 0),(c, 0.67),(d, 0.33),(e, 0)}, $A_{22}$={(a, 0.25),(b, 0.25),(c, 0.5),(d, 0),(e, 0)} and respectively

$x_{c12} = \frac{1}{2}(5 \text{ x } 0.67 + 7 \text{ x } 0.33) = 2.83$, $x_{c22} = \frac{1}{2}(0.25 + 3 \text{ x } 0.25 + 5 \text{ x } 0.25) = 1.125$

By our criterion, the first group again demonstrates a significantly better performance.

Finally, at the third stage of the generation of ideas we have

$A_{13}$= $A_{23}$ = *{(a, 0.25),(b, 0.25),(c, 0.25),(d, 0),(e, 0)},* which obviously means that at this stage the performances of both groups are identical.

Based on our calculations we can conclude that the first group demonstrated a significantly better performance at the stages of imagination and of visualization, but performed identically with the second one at the stage of the generation of ideas.

We have also performed successfully two more couples of experiments with other students giving them for solution the following problems:

*Problem 4:* Let us correspond to each letter the number showing its order into the alphabet (A=1, B=2, C=3 etc). Let us correspond also to each word consisting of 4 letters a 2X2 matrix in the obvious way;

e.g. the matrix $\begin{bmatrix} 19 & 15 \\ 13 & 5 \end{bmatrix}$ corresponds to the word SOME. Using the matrix E=$\begin{bmatrix} 8 & 5 \\ 11 & 7 \end{bmatrix}$ as an encoding matrix how you could send the message LATE in the form of a camouflaged matrix to a receiver knowing the above process and how he (she) could decode your message?

*Problem 5:* A ballot box contains 8 balls numbered from 1 to 8. One makes 3 successive drawings of a lottery, putting back the corresponding ball to the box before the next lottery. Find the probability of getting all the balls that he draws out of the box different.

Due to the lack of space we are not going to present here the results of these experiments, which were obtained in a similar way as above. However, the successful performance of all the experiments shows that our fuzzy model behaves well in practice.

Notice that our model can be also used (in a simplified form) for the individual assessment of the students of each group. In this case a qualitative profile (x, y, z) with x, y, z in *U* is assigned to each student concerning his/her performance at each stage of the reasoning process. This type of assessment by reference to the profile related to each student defines in general a relation o partial order among students' with respect to their total performance. For example, consider the student profiles of Table 1. Then the student possessing the profile (d, b, b) demonstrates a better performance than the student possessing the profile (b, b, b), or (c, b, a), or (d, a, a), etc. However, the student with profile (d, c, a) demonstrates a better performance at the stage of visualization than the student with profile (d, b, b), which demonstrates a better performance at the stage of generation of ideas, etc.

# 5. Conclusions and discussion

The following conclusions can be drawn from the material presented in this paper:

- The main stages of the human reasoning for reaching to a solution of any problem in general involve imagination, visualization and idea generations. The above stages are helpful in understanding the individuals' 'ideal behaviour' during the reasoning process. However, things in real situations are usually not happening like that, since human cognition utilizes in general concepts that are inherently graded and therefore fuzzy.
- In this paper we constructed a fuzzy model for the reasoning process giving, through the calculation of probabilities and possibilities of all possible individuals' profiles, a quantitative/qualitative view of their behaviour during the process. In this model the main stages of human reasoning are represented as fuzzy subsets of set of linguistic labels characterizing a person's performance in each stage.

- Based on the above model and using the coordinates of the centre of gravity of the graph of the corresponding membership function we also developed a method of measuring the reasoning skills of a group of individuals.
- Our model can be also used (in a simplified form) for the individual assessment of the members of each group. In this case a qualitative profile (x, y, z) with x, y, z in *U* is assigned to each member concerning his/her performance at each stage of the reasoning process. This type of assessment defines in general a relation of partial order among the individuals' with respect to their total performance.
- A series of classroom experiments performed with student groups' of T. E. I. of Patras, Greece, shows that our fuzzy model behaves well in practice.

Our plans for future research on the subject involve:

- The possible extension of our fuzzy model for the description of other real life situations involving fuzziness and/or uncertainty.
- Further experimental applications of our model in order to obtain more creditable statistical data.